\documentclass[11pt,a4paper]{article}
\usepackage[hyperref]{acl2019}
\usepackage{times}
\usepackage{latexsym}
\usepackage{multirow}
\usepackage{float}
\usepackage{graphicx}
\usepackage{amsmath}
\usepackage{amsfonts}
\usepackage{amssymb}
\usepackage{todonotes}

\usepackage[utf8]{inputenc}
\usepackage[T1]{fontenc}
\usepackage[russian,english]{babel}
\usepackage{booktabs, tabularx}

\usepackage{url}

\aclfinalcopy 
\def\aclpaperid{712} 

\title{When a Good Translation is Wrong in Context: Context-Aware Machine Translation Improves on Deixis, Ellipsis, and Lexical Cohesion}

\author{Elena Voita$^{1,2}$ \quad Rico Sennrich$^{3,4}$ \quad Ivan Titov$^{3,2}$\bigskip\\
  $^1$Yandex, Russia \quad 
  $^2$University of Amsterdam, Netherlands \\
  $^3$University of Edinburgh, Scotland  \quad
  $^4$University of Zurich, Switzerland \\
  {\tt lena-voita@yandex-team.ru} \\ {\tt rico.sennrich@ed.ac.uk} \quad {\tt ititov@inf.ed.ac.uk}
}

\date{}

\begin{document}
\maketitle
\begin{abstract}

Though machine translation errors caused by the lack of context beyond one sentence have long been acknowledged, the development of context-aware NMT systems is hampered by several problems. Firstly, standard metrics are not sensitive to improvements in consistency in document-level translations. Secondly, previous work on context-aware NMT assumed that the sentence-aligned parallel data consisted of complete documents while in most practical scenarios such document-level data constitutes only a fraction of the available parallel data. To address the first issue, we perform a human study on an English-Russian subtitles dataset and identify deixis, ellipsis and lexical cohesion as three main sources of inconsistency. We then create test sets targeting these phenomena. To address the second shortcoming, we consider a set-up in which a much larger amount of sentence-level data is available compared to that aligned at the document level. We introduce a model that is suitable for this scenario and demonstrate major gains over a context-agnostic baseline on our new benchmarks without sacrificing performance as
measured with BLEU.\footnote{We release code and data sets at \url{https://github.com/lena-voita/good-translation-wrong-in-context}.}

\end{abstract}

\section{Introduction }

With the recent rapid progress of neural machine translation (NMT), translation mistakes and inconsistencies due to the lack of extra-sentential context are becoming more and more noticeable among otherwise adequate translations produced by standard context-agnostic NMT systems \cite{D18-1512}. Though this problem has
recently triggered a lot of attention to context-aware translation~\cite{jean_does_2017,wang_exploiting_2017,tiedemann_neural_2017,bawden2017,voita18,maruf-haffari:2018:Long,agrawal2018,miculicich-EtAl:2018:EMNLP,zhang-etal-2018-improving}, the progress and wide-spread adoption of the new paradigm is hampered by several important problems. Firstly, it is highly non-trivial to design metrics which would reliably trace the progress and guide model design. Standard machine translation metrics (e.g., BLEU) do not appear appropriate as they do not sufficiently differentiate between consistent and inconsistent translations~\cite{wong-kit:2012:EMNLP-CoNLL}.\footnote{We use the term `inconsistency' to refer to any violations causing good translations of isolated sentences not to work together, independently of which linguistic phenomena (e.g., ellipsis or lexical cohesion) impose the violated constraints.}
For example, if multiple translations of a name are possible, forcing consistency is essentially as likely to make all occurrences of the name match the reference translation as making them all different from the reference.
Second, most previous work on context-aware NMT has made the assumption that all the bilingual data is available at the document level. 
However, isolated parallel sentences are a lot easier to acquire and hence only a fraction of the parallel data will be at the document level in any practical scenario. In other words, a context-aware model trained only on document-level parallel data is highly unlikely to outperform a context-agnostic model estimated from much larger sentence-level parallel corpus. This work aims to address both these shortcomings.

A context-agnostic NMT system would often produce plausible translations of isolated sentences, however, when put together in a document, these translations end up being inconsistent with each other.
We investigate which linguistic phenomena cause the inconsistencies using the OpenSubtitles~\cite{LISON18.294} corpus for the English-Russian language pair.
We identify deixis, ellipsis and lexical cohesion as three main sources of the violations, together amounting to about 80\% of the cases.
We create test sets focusing specifically on the three identified phenomena (6000 examples in total).

We show that by using a limited amount of document-level parallel data, we can already achieve substantial improvements on these benchmarks without negatively affecting performance as measured with BLEU. 
Our approach is inspired by the Deliberation Networks~\cite{deliberation-networks}. In our method, the initial translation produced by a baseline context-agnostic model is refined by a context-aware system which is trained on a small document-level subset of parallel data. 

The key contributions are as follows:
\begin{itemize}
    \item we analyze which phenomena cause context-agnostic translations to be inconsistent with each other;
    \item we create test sets specifically addressing the most frequent phenomena;
    \item we consider a novel and realistic set-up where a much larger amount of sentence-level data is available compared to that aligned at the document level;
    \item we introduce a model suitable for this scenario, and demonstrate that it is effective on our new benchmarks without sacrificing performance as measured with BLEU.
\end{itemize}

\section{Analysis}
\label{sect:analysis}

We begin with a human study, in which we:
\begin{enumerate}
    \item identify cases when good sentence-level translations are not good when placed in context of each other,
    \item categorize these examples according to the phenomena leading to a discrepancy in translations of consecutive sentences.
\end{enumerate}
The test sets introduced in Section~\ref{sec:all_testsets} will then target the most frequent phenomena.

\subsection{Human annotation}
To find what makes good context-agnostic translations incorrect when placed in context of each other, we start with pairs of consecutive sentences. We gather data with context from the publicly available OpenSubtitles2018 corpus~\cite{LISON18.294} for English and Russian. We train a context-agnostic Transformer on 6m sentence pairs. Then we translate 2000 pairs of consecutive sentences using this model. For more details on model training and data preprocessing, see Section~\ref{sect:data_setting}.

Then we use human annotation to assess the adequacy of the translations without context and in the context of each other.
The whole process is two-stage:
\begin{enumerate}
\item sentence-level evaluation: we ask if the translation of a given sentence is good,
\item evaluation in context: for pairs of consecutive good translations according to the first stage, we ask if the translations are good in context of each other.
\end{enumerate}

In the first stage, the annotators are instructed to mark as ``good'' translations which (i) are fluent sentences in the target language (in our case, Russian) (ii) can be reasonable translations of a source sentence in some context. 

For the second stage we only consider pairs of sentences with good sentence-level translations. The annotators are instructed to mark translations as bad in context of each other only if there is no other possible interpretation or extra additional context which could have made them appropriate. This was made to get more robust results, avoiding the influence of personal preferences of the annotators (for example, for using formal or informal speech), and excluding ambiguous cases that can only be resolved with additional context.

\begin{table}[t!]
\centering
\begin{center}
\begin{tabular}{|c|c|cc|}
\hline
\multirow{2}{*}{\bf all}  & 
  \multirow{2}{*}{\bf one/both bad} & \multicolumn{2}{c|}{\bf both good}\\
& \bf & \bf bad pair & \bf good pair \\
\hline
2000 & 211 & 140 & 1649 \\ \hline
100\% & 11\% & 7\% & 82\% \\
\hline
\end{tabular}
\end{center}
\vspace{-1ex}
\caption{\label{tab:annotation_general_stat} Human annotation statistics of pairs of consecutive translation.}
\vspace{-2ex}
\end{table}

The statistics of answers are provided in Table~\ref{tab:annotation_general_stat}.
We find that our annotators labelled $82\%$ of sentence pairs as good translations. In $11\%$ of cases, at least one translation was considered bad at the sentence level, and in another $7\%$, the sentences were considered individually good, but bad in context of each other. This indicates that in our setting, a substantial proportion of translation errors are only recognized as such in context.

\subsection{Types of phenomena}
\label{sec:phenomena_types}
From the results of the human annotation, we take all instances of consecutive sentences with good translations which become incorrect when placed in the context of each other. For each, we identify the language phenomenon which caused a discrepancy. The results are provided in Table~\ref{tab:annotation_general_types}.

Below we discuss these types of phenomena, as well as problems in translation they cause, in more detail. In the scope of current work, we concentrate only on the three most frequent phenomena.

\begin{table}[t!]
\begin{center}
\begin{tabular}{|l|c|}
\hline \bf \bf type of phenomena  & \bf  frequency \\ \hline
deixis & 37\%  \\ 
ellipsis & 29\% \\
lexical cohesion & 14\% \\
ambiguity & \phantom{0}9\% \\
anaphora & \phantom{0}6\% \\
other & \phantom{0}5\% \\
\hline
\end{tabular}
\end{center}
\vspace{-1ex}
\caption{\label{tab:annotation_general_types} Types of phenomena causing discrepancy in context-agnostic translation of consecutive sentences when placed in the context of each other}
\vspace{-1ex}
\end{table}

\subsubsection{Deixis}
\label{sec:phenomena_deixis}
In this category, we group several types of deictic words or phrases, i.e. referential expressions whose denotation depends on context. This includes personal deixis (``I'', ``you''), place deixis (``here'', ``there''), and discourse deixis, where parts of the discourse are referenced (``that's a good question.''). Most errors in our annotated corpus are related to person deixis, specifically gender marking in the Russian translation, and the T-V distinction between informal and formal you (Latin ``tu'' and ``vos''). 

\begin{table}[t!]
\begin{center}
\begin{tabular}{|l|c|}
\hline \bf \bf type of discrepancy  & \bf frequency \\ \hline
T-V distinction & 67\%   \\ 
speaker/addressee gender: & \\
\hspace{1cm} same speaker & 22\%  \\
\hspace{1cm} different speaker & \phantom{0}9\%  \\
other & 
\phantom{0}2\%  \\
\hline
\end{tabular}
\end{center}
\vspace{-1ex}
\caption{\label{tab:deixis_disc_types} Types of discrepancy in context-agnostic translation caused by deixis (excluding anaphora)}
\vspace{-2ex}
\end{table}

\begin{figure}[t!]
\center{\includegraphics[scale=0.38]{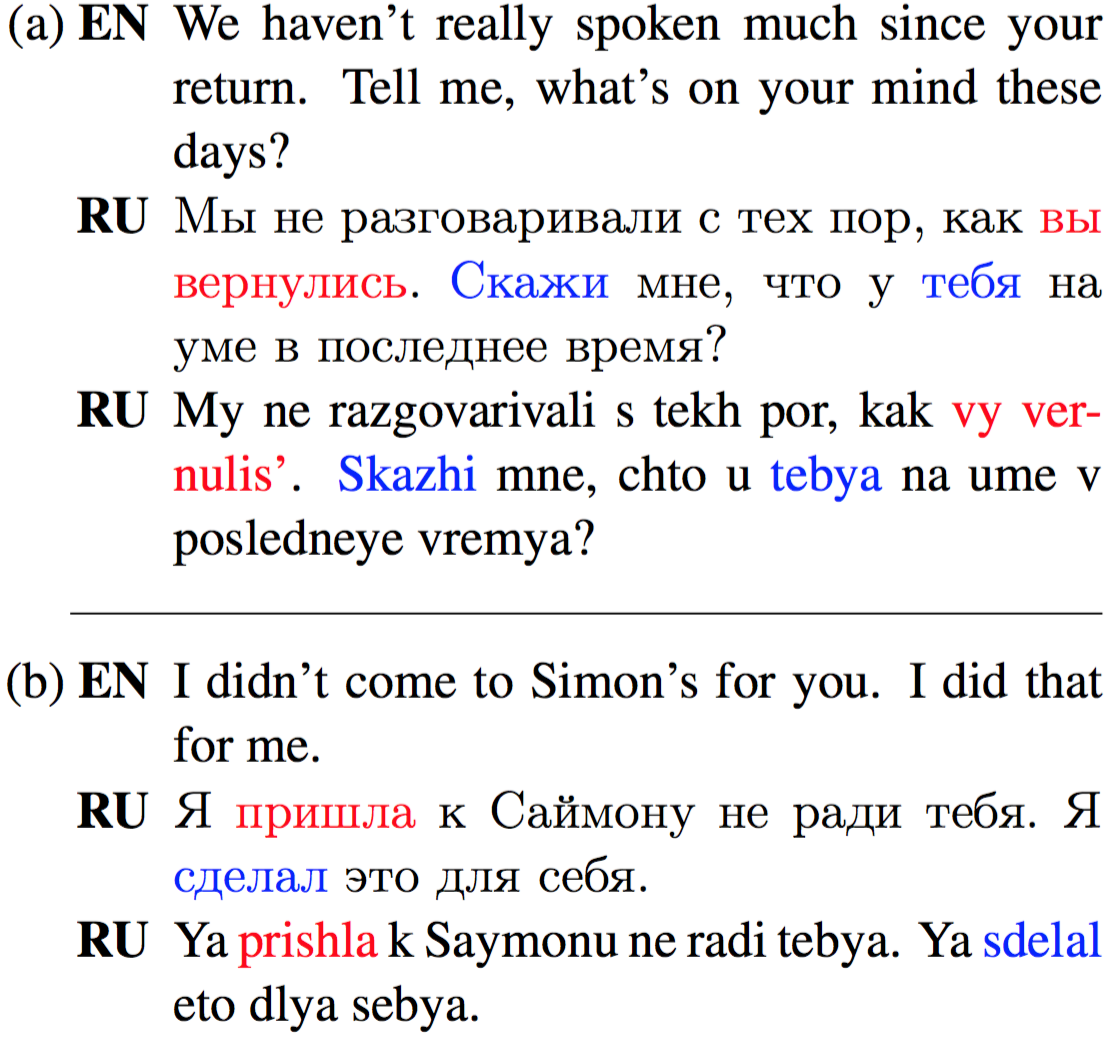}}
\caption{Examples of violation of (a) T-V form consistency, (b) speaker gender consistency.
\newline In color: (a) red -- V-form, blue -- T-form; (b) red -- feminine, blue -- masculine.} \label{fig:deixis_examples}
\vspace{-2ex}
\end{figure}

In many cases, even when having access to neighboring sentences,
one cannot make a confident decision which of the
forms should be used, as there are no obvious markers
pointing to one form or another  (e.g., for the T-V distinction, words such as ``officer'', ``mister'' for formal and ``honey'', ``dude'' for informal). 
However, when 
pronouns refer to the same person, the pronouns, as well as verbs that agree with them, should be translated using the same form.
See Figure~\ref{fig:deixis_examples}(a) for an example translation that violates T-V consistency.
Figure~\ref{fig:deixis_examples}(b) shows an example of inconsistent first person gender (marked on the verb), although the speaker is clearly the same.

Anaphora are a form of deixis that received a lot of attention in MT research, both from the perspective of modelling~\cite[among others]{lenagard-koehn:2010:WMT, hardmeier2010,W17-4806,bawden2017,voita18} and targeted evaluation ~\cite{W15-2501, GUILLOU16.327, muller-EtAl:2018:WMT}, and we list anaphora errors separately, and will not further focus on them.

\subsubsection{Ellipsis}
\label{sec:phenomena_ellipsis}
Ellipsis is the omission from a clause of one or more words that are nevertheless understood in the context of the remaining elements. 

\begin{figure}[t!]
\center{\includegraphics[scale=0.30]{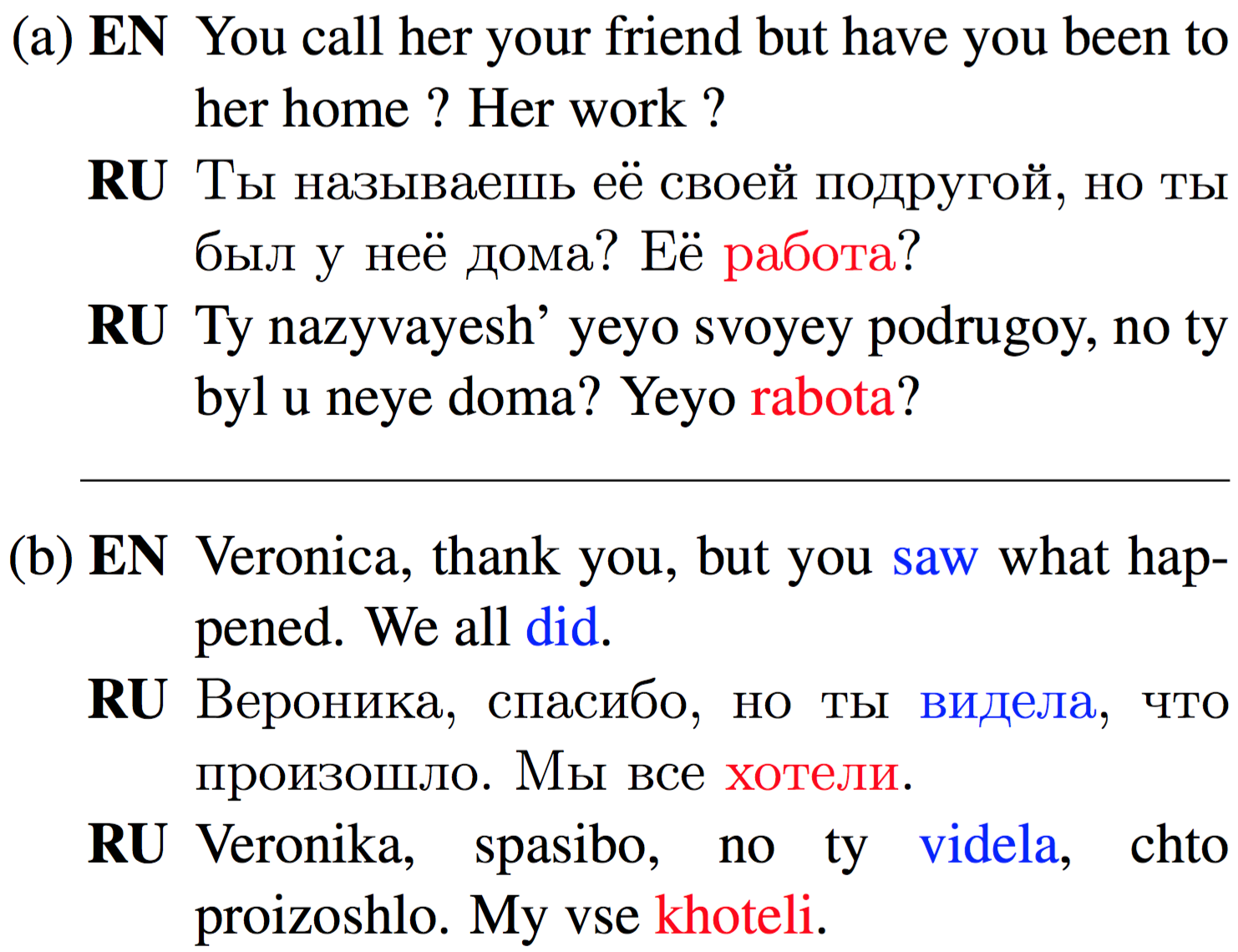}}
\caption{Examples of discrepancies caused by ellipsis. (a) wrong morphological form, incorrectly marking the noun phrase as a subject. (b) correct meaning is ``see'', but MT produces \begin{otherlanguage}{russian}\emph{хотели} \end{otherlanguage} \emph{khoteli} (``want'').} \label{fig:ellipsis_examples}
\vspace{-2ex}
\end{figure}

In machine translation, elliptical constructions in the source language pose a problem if the target language does not allow the same types of ellipsis (requiring the elided material to be predicted from context), or if the elided material affects the syntax of the sentence; for example, the grammatical function of a noun phrase and thus its inflection in Russian may depend on the elided verb (Figure~\ref{fig:ellipsis_examples}(a)), or the verb inflection may depend on the elided subject. Our analysis focuses on ellipses that can only be understood and translated with context beyond the sentence-level. This has not been studied extensively in MT research.\footnote{Exceptions include \citep{P98-2233}, and work on the related phenomenon of pronoun dropping \citep{russo-loaiciga-gulati:2012:SRWEACL2012,wang-EtAl:2016:N16-13,rios-tuggener:2017:EACLshort}.}

We classified ellipsis examples which lead to errors in sentence-level translations by the type of error they cause. Results are provided in Table~\ref{tab:ellipsis_types}.

\begin{table}[t!]
\begin{center}
\begin{tabular}{|l|c|}
\hline \bf \bf type of discrepancy  & \bf frequency \\ \hline
wrong morphological form & 66\%  \\
wrong verb (VP-ellipsis) & 20\% \\
other error  & 14\% \\
\hline
\end{tabular}
\end{center}
\caption{\label{tab:ellipsis_types} Types of discrepancy in context-agnostic translation caused by ellipsis }
\vspace{-1ex}
\end{table}

It can be seen that the most frequent problems related to ellipsis that we find in our annotated corpus are wrong morphological forms, followed by wrongly predicted verbs in case of verb phrase ellipsis in English, which does not exist in Russian, thus requiring the prediction of the verb in the Russian translation (Figure~\ref{fig:ellipsis_examples}(b)).

\subsubsection{Lexical cohesion}
\label{sec:phenomena_consistency}
Lexical cohesion has been studied previously in MT~\cite[among others]{tiedemann:2010:DANLP,gong-zhang-zhou:2011:EMNLP,wong-kit:2012:EMNLP-CoNLL,C18-1050,miculicich-EtAl:2018:EMNLP}.

\begin{figure}[t!]
\center{\includegraphics[scale=0.30]{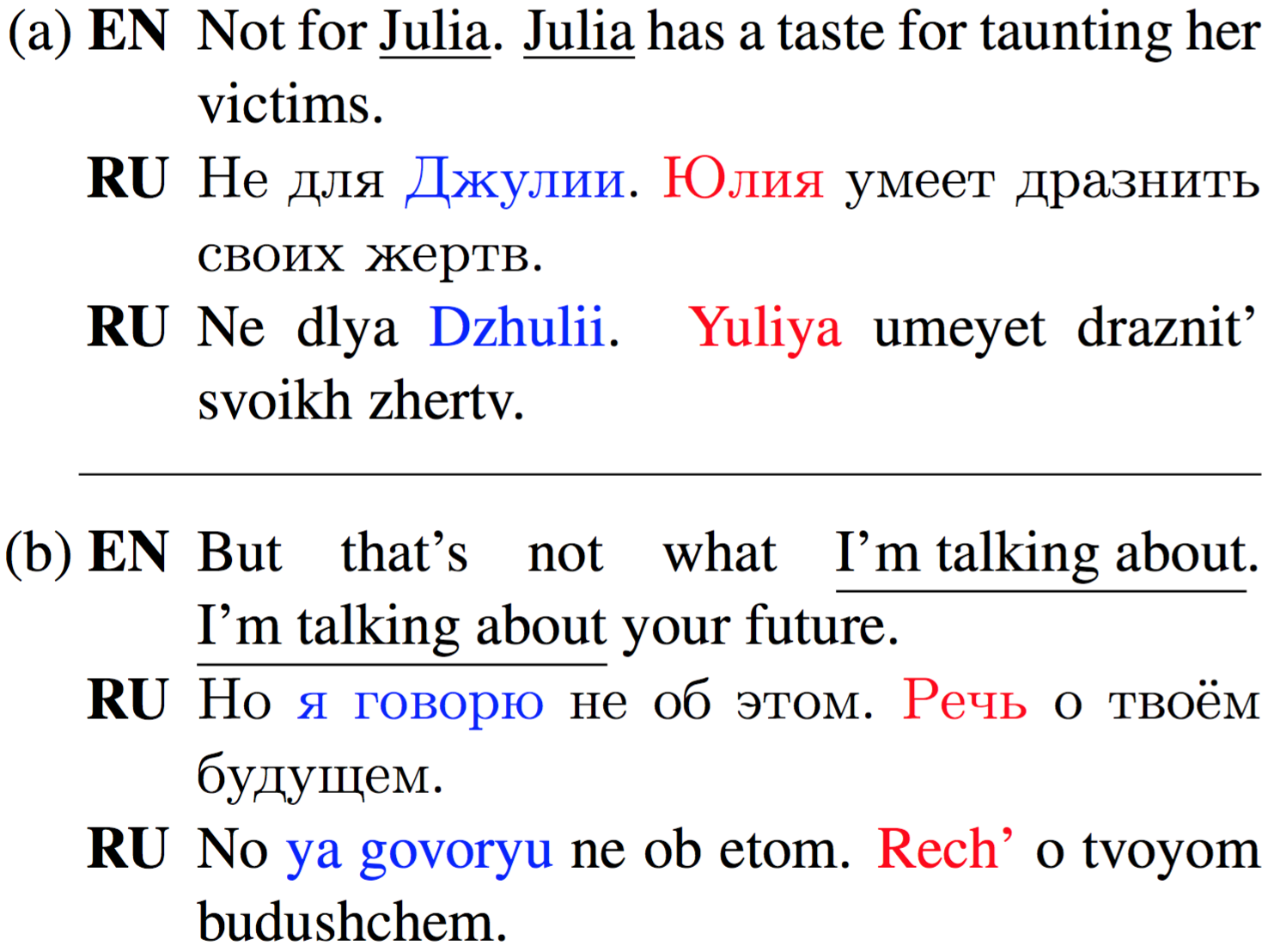}}
\caption{Examples of lack of lexical cohesion in MT. (a) Name translation inconsistency. (b) Inconsistent translation. Using either of the highlighted translations consistently would be good.} \label{fig:consistency_examples}
\vspace{-2ex}
\end{figure}

There are various cohesion devices~\cite{morris-hirst}, and a good translation should exhibit lexical cohesion beyond the sentence level.
 We focus on repetition with two frequent cases in our annotated corpus being reiteration of named entities (Figure~\ref{fig:consistency_examples}(a)) and reiteration of more general phrase types for emphasis (Figure~\ref{fig:consistency_examples}(b)) or in clarification questions.

\section{Test Sets}
\label{sec:all_testsets}
For the most frequent phenomena from the above analysis we create test sets for targeted evaluation. 

Each test set contains contrastive examples. It is specifically designed to test the ability of a system to adapt to contextual information and handle the phenomenon under consideration. Each test instance consists of a true example (sequence of sentences and their reference translation from the data) and several contrastive translations which differ from the true one only in the considered aspect. All contrastive translations we use are correct plausible translations at a sentence level, and only context reveals the errors we introduce. All the test sets are guaranteed to have the necessary context in the provided sequence of 3 sentences. The system is asked to score each candidate example, and we compute the system accuracy as the proportion of times the true translation is preferred over the contrastive ones.

Test set statistics are shown in Table~\ref{tab:deixis_consistency_testset_stat}. 

\subsection{Deixis}
\label{sec:deixis_testsets}
From Table~\ref{tab:deixis_disc_types}, we see that the most frequent error category related to deixis in our annotated corpus is the inconsistency of T-V forms when translating second person pronouns. The test set we construct for this category tests the ability of a 
machine translation system to produce translations with consistent level of politeness. 

We semi-automatically identify sets of consecutive sentences with consistent politeness markers on pronouns and verbs (but without nominal markers such as ``'Mr.'' or ``officer'') and switch T and V forms. Each automatic step was followed by human postprocessing, which ensures the quality of the final test sets.\footnote{Details are provided in the appendix.}
This gives us two sets of translations for each example, one consistently informal (T), and one consistently formal (V). For each, we create an inconsistent contrastive example by switching  the formality of the last sentence. The symmetry of the test set ensures that any context-agnostic model has 50\% accuracy on the test set. 

\begin{table}[t!]
\centering
\begin{tabular}{lcccc}
\toprule
& & \multicolumn{3}{c}{latest relevant context}\\
  & total & 1st & 2nd & 3rd \\ 
 \cmidrule{2-5}
\bf deixis   & 3000 & 1000 & 1000 & 1000\\
\bf lex. cohesion   & 2000 & \phantom{0}855 & \phantom{0}630 & \phantom{0}515\\
\bf ellipsis (infl.) & \phantom{0}500\\
\bf ellipsis (VP) & \phantom{0}500\\
\bottomrule
\end{tabular}\textbf{}
\caption{Size of test sets: total number of test instances and with regard to the latest context sentence with politeness indication or with the named entity under consideration. For ellipsis, we distinguish whether model has to predict correct noun phrase inflection, or correct verb sense (VP ellipsis).}\label{tab:deixis_consistency_testset_stat}
\vspace{-2ex}
\end{table}

\subsection{Ellipsis}
\label{sec:ellipsis_testsets}
From Table~\ref{tab:ellipsis_types}, we see that the two most frequent types of ambiguity caused by the presence of an elliptical structure have different nature, hence we construct individual test sets for each of them.

Ambiguity of the first type comes from the inability to predict the correct morphological form of some words. We manually gather examples with such structures in a source sentence and change the morphological inflection of the relevant target phrase to create contrastive translation.
Specifically, we focus on noun phrases where the verb is elided, and the ambiguity lies in how the noun phrase is inflected. 

The second type we evaluate are verb phrase ellipses. Mostly these are sentences with an auxiliary verb ``do'' and omitted main verb. We manually gather such examples and replace the translation of the verb, which is only present on the target side, with other verbs with different meaning, but the same inflection. Verbs which are used to construct such contrastive translations are the top-10 lemmas of translations of the verb ``do'' which we get from the lexical table of Moses~\cite{moses} induced from the training data.

\subsection{Lexical cohesion}
\label{sec:consistency_testsets}
Lexical cohesion can be established for various types of phrases and can involve reiteration or other semantic relations. In the scope of the current work, we focus on the reiteration of entities, since these tend to be non-coincidental, and can be easily detected and transformed.

We 
identify named entities with alternative translations into Russian, find passages where they are translated consistently, and create contrastive test examples by switching the translation of some instances of the named entity. For more details, please refer to the appendix.

\section{Model and Setting}
\label{sect:model}

\subsection{Setting}

Previous work on context-aware neural machine translation used data where all training instances have context. This setting limits the set of available training sets one can use: in a typical scenario, we have a lot of sentence-level parallel data and only a small fraction of document-level data. Since machine translation quality depends heavily on the amount of training data,  training a context-aware model is counterproductive if this leads to ignoring the majority of available sentence-level data and sacrificing general quality. 
We will also show that a naive approach to combining sentence-level and document-level data leads to a drop in performance. 

In this work, we argue that it is important to consider an asymmetric setting where the amount of available document-level data is much smaller than that of sentence-level  data, and propose an approach specifically targeting this scenario.

\subsection{Model}

\begin{figure*}[t!]
\center{\includegraphics[scale=0.35]{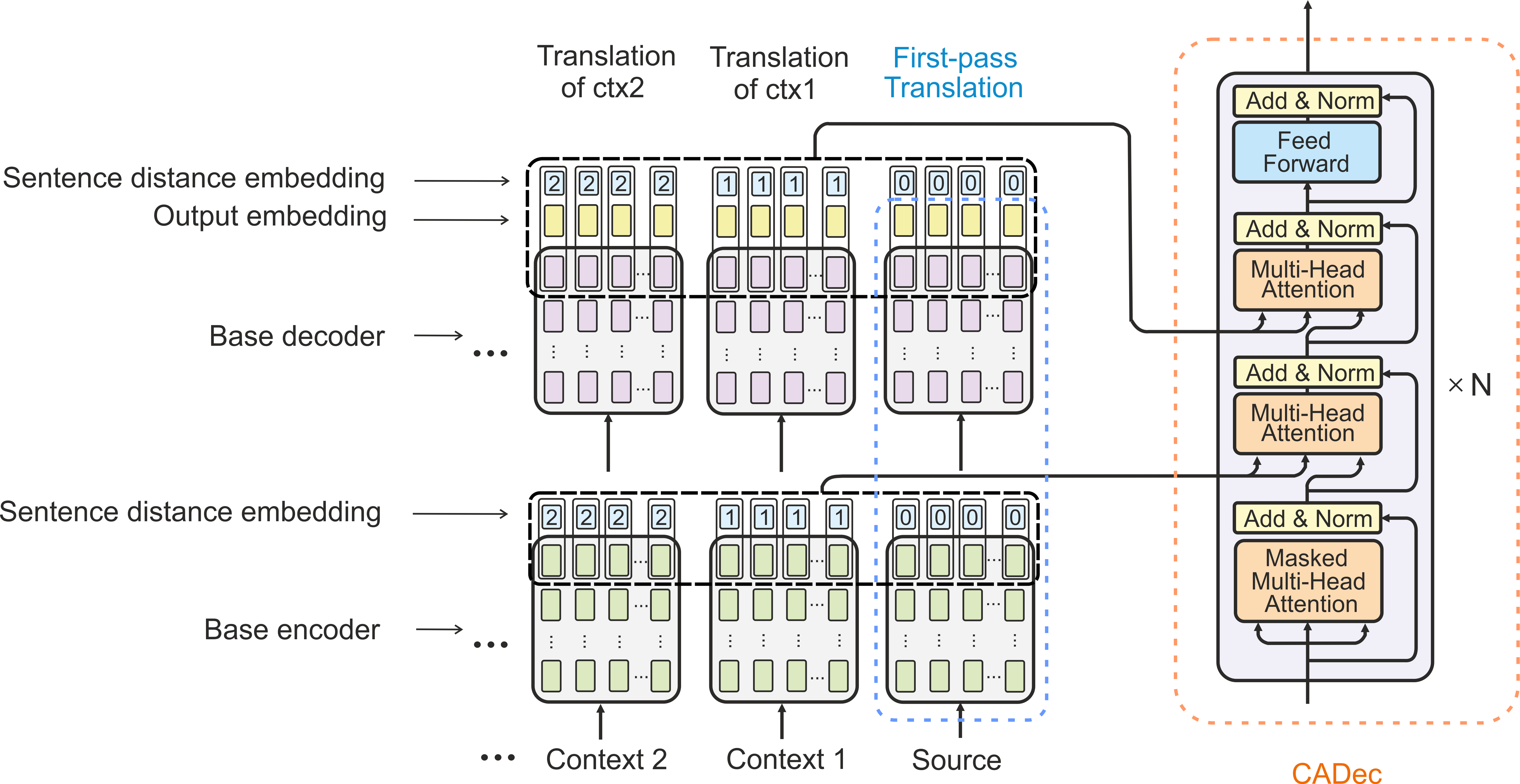}}
\vspace{-1ex}
\caption{Model architecture}
\vspace{-1ex}
\label{fig:model_arcitecture}
\end{figure*}

We introduce a two-pass framework: first, the sentence is translated with a context-agnostic model, and then this translation is refined using context of several previous sentences (context includes source sentences as well as  their translations). 
We expect this architecture to be suitable in the proposed setting: the baseline context-agnostic model can be trained on a large amount of sentence-level data, and the second-pass model can be estimated on a smaller subset of parallel data which includes context. As the first-pass translation is produced by a strong model, we expect no loss in general performance when training the second part on a smaller dataset.

The model is close in spirit to the Deliberation networks~\cite{deliberation-networks}. 
The first part of the model is a context-agnostic model (we refer to it as the base model), and the second one is a context-aware decoder (CADec) which refines context-agnostic translations using context. The base model is trained on sentence-level data and then fixed. It is used only to sample context-agnostic translations and to get vector representations of the source and translated sentences. CADec is trained only on data with context.

Let $D_{sent}=\{(x_{i}, y_{i})\}_{i=1}^{N}$ denote the sentence-level data with $n$ paired sentences and $D_{doc}=\{(x_{j}, y_{j}, c_{j})\}_{j=1}^{M}$ denote the 
document-level data, where $(x_j, y_j)$ is source and target sides of a sentence to be translated, $c_j$ are several preceding sentences along with their translations. 

\textbf{Base model} For the baseline context-agnostic model we use the original Transformer-base~\cite{attention-is-all-you-need}, trained to maximize the sentence-level  log-likelihood $\frac{1}{N}\sum\limits_{(x_i, y_i)\in D_{sent}}\log P(y_i|x_i, \theta_B)$.

\textbf{Context-aware decoder (CADec)}
The context-aware decoder is trained to correct translations given by the base model using contextual information. Namely, we maximize the following document-level log-likelihood: 
\begin{align}
\nonumber
{\frac{1}{M}} \!\!\! \sum_{(x_j, y_j)\in D_{doc}}{\!\!\!\!\!\!\!\!\! \log
E_{y^B_j \propto P(y|x_j, \theta_B)} 
     P(y_j|x_j, y^B_j, c_j, \theta_{C})
},
\end{align}
where $y^B_j$ is sampled from  $P(y|x_j, \theta_B)$.

CADec is composed of a stack of $N=6$ identical layers and is similar to the decoder of the original Transformer.  It has a masked self-attention layer and attention to encoder outputs, and additionally each layer has a block attending over the outputs of the base decoder~(Figure~\ref{fig:model_arcitecture}). 
We 
use the states from the last layer of the base model's encoder of the current source sentence and all context sentences as input to the first multi-head attention. For the second multi-head attention we input both last states of the base decoder and the target-side token embedding layer; this is done for translations of the source and also all context sentences. All sentence representations are produced by the base model.
To encode the relative position of each sentence, we concatenate both the encoder and decoder states with one-hot vectors representing their position (0 for the source sentence,
1 for the immediately preceding one, etc). These distance embeddings are shown in blue in Figure~\ref{fig:model_arcitecture}.

\section{Experiments}
\label{sect:experiments}

\subsection{Training}
\label{sec:training}
At training time, we use reference translations as translations of the previous sentences.
For the current sentence, we either sample a translation from the base model or use a corrupted version of the reference translation. We propose to stochastically mix objectives corresponding to these versions: 
\begin{multline*}
\frac{1}{M}\sum\limits_{(x_j, y_j)\in D_{doc}}\log\Big[b_j\cdot P(y_j|x_j,\tilde{y_{j}}, c_j, \theta_{C})) +\\
+(1-b_j)\cdot P(y_j|x_j,y^B_j, c_j, \theta_{C})\Big],
\end{multline*}
where $\tilde{y_j}$ is a corrupted version of the reference translation and $b_j \in \{0,1\}$ is drawn from Bernoulli distribution with parameter $p$,  $p=0{.}5$ in our experiments.  
Reference translations are corrupted by replacing 20\% of their tokens with random tokens. 

We discuss the importance of the proposed training strategy, as well as the effect of varying the value of $p$, in Section~\ref{sect:ablation_noise_reference}.

\subsection{Inference}
\label{sec:inference}
As input to CADec for the current sentence, we use the translation produced by the base model. Target sides of the previous sentences are produced
by our two-stage approach for those sentences which have context and with the base model for those which do not. 
We use beam search with a beam of 4 for 
all models.

\subsection{Data and setting}
\label{sect:data_setting}
We use the publicly available OpenSubtitles2018 corpus~\cite{LISON18.294} for English and Russian. As described in detail in the appendix, we apply data cleaning after which only a fraction of data has context of several previous sentences. We use up to 3 context sentences in this work. We randomly choose 6 million training instances from the resulting data, among which 1{.}5m have context of three sentences. We randomly choose two subsets of 10k instances for development and testing and construct our contrastive test sets from 400k held-out instances from movies not encountered in training. 
The hyperparameters, preprocessing and training details are provided in the supplementary material.

\section{Results}
We evaluate in two different ways: using 
BLEU for general quality and the proposed contrastive test sets for consistency. We show that models indistinguishable with BLEU can be very different in terms of consistency.

We randomly choose 500 out of 2000 examples from the lexical cohesion set and 500 out of 3000 from the deixis test set for validation and leave the rest for final testing. 
We compute BLEU  on the development set as well as scores on lexical cohesion and deixis development sets. We use convergence in both metrics to decide when to stop training.
The importance of using both criteria is discussed in Section~\ref{sect:context_aware_stoppoing_criteria}.
After the convergence, we average~5 checkpoints and report scores on the final test sets.

\subsection{Baselines}

We consider three baselines. 

\textbf{baseline} The context-agnostic baseline is Transformer-base trained on all sentence-level data. Recall that it is also used as the base model in our 2-stage approach. 

\textbf{concat} The first context-aware baseline is a simple concatenation model. It is trained on 6m sentence pairs, including 1{.}5m having 3 context sentences. For the concatenation baseline, we use a special token separating sentences (both on the source and target side). 

\textbf{s-hier-to-2.tied} This is the version of the model \textit{s-hier-to-2} introduced by~\citet{bawden2017}, where
the parameters between encoders are shared~\cite{muller-EtAl:2018:WMT}. The model has an additional encoder for source context, whereas the target side of the corpus is concatenated, in the same way as for the concatenation baseline. 
Since the model is suitable only for one context sentence, it is trained on 6m sentence pairs, including 1{.}5m having one context sentence. We chose \textit{s-hier-to-2.tied} as our second context-aware baseline because it also uses context on the target side and performed best in a contrastive evaluation of pronoun translation ~\cite{muller-EtAl:2018:WMT}.

\subsection{General results}
BLEU scores for our model and the baselines are given in Table~\ref{tab:bleu_scores}.\footnote{We use bootstrap resampling \cite{koehn2004statistical} for significance testing.} For context-aware models, all sentences in a group were translated, and then only the current sentence is evaluated. We also report BLEU for the context-agnostic baseline trained only on 1{.}5m dataset to show how the performance is influenced by the amount of data.

\begin{table}[t!]
\centering
\begin{tabular}{lc}
\toprule
 \bf model & \bf BLEU \\ 
 \cmidrule{1-2}
baseline (1{.}5m)   &  29{.}10 \\
baseline (6m)   &  \bf 32{.}40 \\
concat   & 31{.}56\\
s-hier-to-2.tied & 26{.}68 \\
CADec & \bf 32{.}38\\
\bottomrule
\end{tabular}\textbf{}
\caption{BLEU scores. CADec trained with $p=0{.}5$.  Scores for CADec are not statistically different from the baseline (6m).}
\label{tab:bleu_scores}
\vspace{-2ex}
\end{table}

We observe that our model is no worse in BLEU than the baseline despite the second-pass model being trained only on a fraction of the data. In contrast, the concatenation baseline, trained on a mixture of data with and without context  
is about 1 BLEU below the context-agnostic baseline and our model when using all 3 context sentences. CADec's performance remains the same independently from the number of context sentences (1, 2 or~3) as measured with BLEU.

\textit{s-hier-to-2.tied} performs worst in terms of BLEU, but note that this is a shallow recurrent model, while others are Transformer-based. It also suffers from the asymmetric data setting, like the concatenation baseline.

\subsection{Consistency results}

Scores on the deixis, cohesion and ellipsis test sets are provided in Tables~\ref{tab:deixis_consistency_scores} and \ref{tab:ellipsis_scores}. For all tasks, we observe a large improvement from using context. 
For deixis, the concatenation model (concat) and CADec improve over the baseline by 33.5 and 31.6 percentage points, respectively. 
On the lexical cohesion test set, CADec shows a large improvement over the context-agnostic baseline (12.2 percentage points), while concat performs similarly to the baseline. For ellipsis, both models improve substantially over the baseline (by 19-51 percentage points), with concat stronger for inflection tasks and CADec stronger for VP-ellipsis. Despite its low BLEU score, \textit{s-hier-to-2.tied} also shows clear improvements over the context-agnostic baseline in terms of consistency, but underperforms both the concatenation model and CADec, which is unsurprising given that it uses only one context sentence. When looking only at the scores where the latest relevant context is in the model's context window (column 2 in Table~\ref{tab:deixis_consistency_scores}), \textit{s-hier-to-2.tied} outperforms the concatenation baseline for lexical cohesion, but remains behind the performance of CADec.

The proposed test sets let us distinguish models which are otherwise identical in terms of BLEU: the performance of the baseline and CADec is the same when measured with BLEU, but very different in terms of handling contextual phenomena.

\begin{table}[t!]
\centering
\begin{tabular}{lcccc}
\toprule
&   & \multicolumn{3}{c}{latest relevant context}\\
& total & 1st & 2nd & 3rd \\ 
 \cmidrule{1-5}
\multicolumn{3}{l}{\bf deixis}\\
baseline   &  50{.}0 & 50{.}0 & 50{.}0 & 50{.}0\\
concat   & \bf 83{.}5 & \bf 88{.}8  & \bf 85{.}6 & \bf 76{.}4\\
s-hier-to-2.tied & 60{.}9 & 83{.}0 & 50{.}1 & 50{.}0 \\
CADec 
& 81{.}6 & 84{.}6 & 84{.}4 & 75{.}9\\
\cmidrule{1-5}
\multicolumn{3}{l}{\bf lexical cohesion}\\
baseline   &  45{.}9 & 46{.}1 & 45{.}9 & 45{.}4\\
concat   & 47{.}5 & 48{.}6  & 46{.}7 & 46{.}7\\
s-hier-to-2.tied & 48{.}9 & 53{.}0 & 46{.}1 & 45{.}4 \\
CADec 
& \bf 58{.}1 & \bf 63{.}2 & \bf 52{.}0 & \bf 56{.}7\\
\bottomrule
\end{tabular}\textbf{}
\caption{Accuracy for deixis and lexical cohesion.} \label{tab:deixis_consistency_scores}
\end{table}

\begin{table}[t!]
\centering
\begin{tabular}{lcc}
\toprule
  & {\bf ellipsis (infl.)} & {\bf ellipsis (VP)}\\ 
 \cmidrule{1-3}
baseline & 53{.}0 & 28{.}4\\
concat & \bf 76{.}2 & 76{.}6\\
s-hier-to-2.tied & 66{.}4 & 65{.}6 \\
CADec 
& 72{.}2 & \bf 80{.}0\\

\bottomrule
\end{tabular}\textbf{}
\caption{Accuracy on ellipsis test set.}\label{tab:ellipsis_scores}
\end{table}

\subsection{Context-aware stopping criteria}
\label{sect:context_aware_stoppoing_criteria}

\begin{figure}[t!]
\center\includegraphics[scale=0.13]{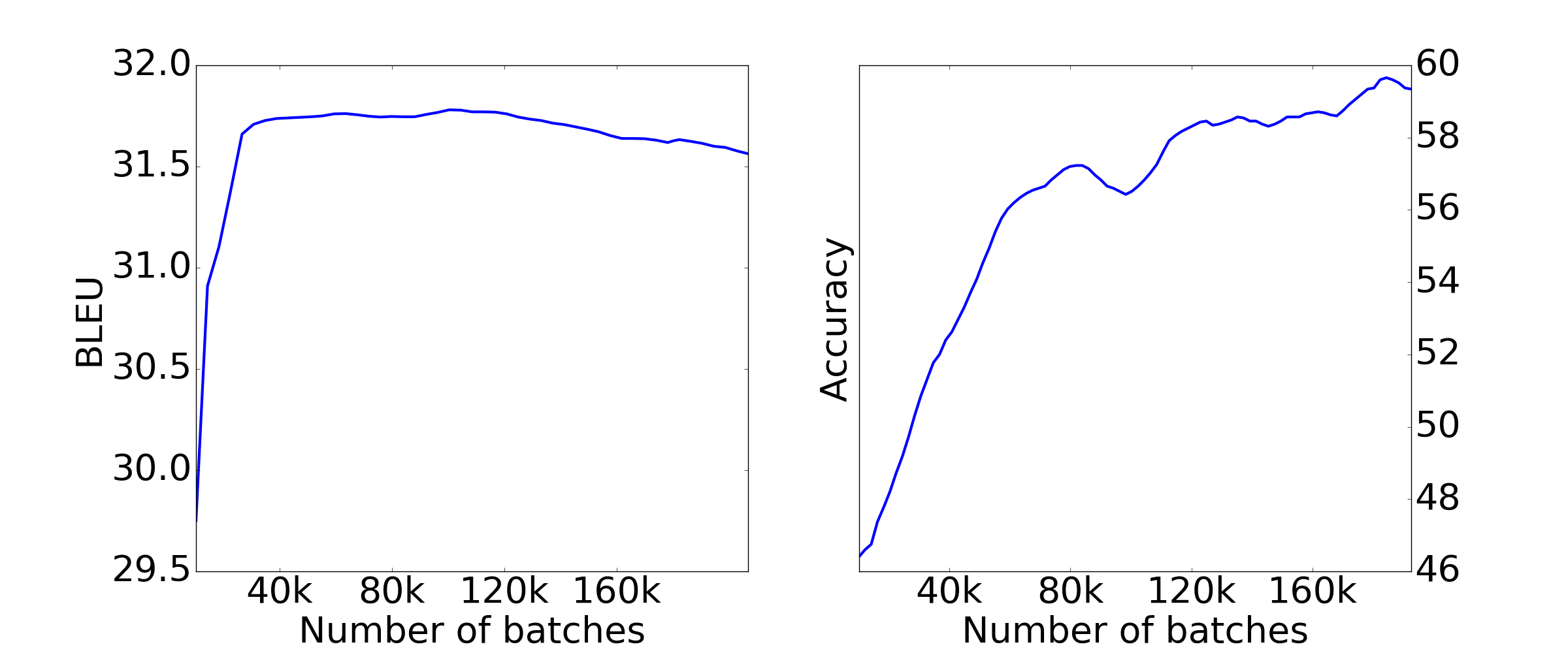}
\vspace{-2ex}
\caption{BLEU and lexical cohesion accuracy on the development sets during CADec training.}
\label{fig:bleu}
\end{figure}

Figure~\ref{fig:bleu} shows that for context-aware models, BLEU is not sufficient as a criterion for stopping: even when a model has converged in terms of BLEU, it continues to improve in terms of consistency. For CADec trained with $p=0{.}5$, BLEU score has stabilized after 40k batches, but the lexical cohesion score continues to grow.

\subsection{Ablation: using corrupted reference}
\label{sect:ablation_noise_reference}

\begin{table}[t!]
\centering
\begin{tabular}{lcccc}
\toprule
 $p$ & \bf BLEU & \bf deixis & \bf lex. c. & {\bf ellipsis} \\ 
 \cmidrule{1-5}
$p\!=\!0$ & 32{.}34 & 84{.}1 & 48{.}7 & 65 / 75\\
$p\!=\!0{.}25$ & 32{.}31 & 83{.}3 & 52{.}4 & 67 / 78\\
$p\!=\!0{.}5$ & 32{.}38 & 81{.}6 & 58{.}1 & 72 / 80\\
$p\!=\!0{.}75$ & 32{.}45 & 80{.}0 & 65{.}0 & 70 / 80\\
\bottomrule
\end{tabular}\textbf{}
\vspace{-1ex}
\caption{Results for different probabilities of using corrupted reference at training time. BLEU for 3 context sentences.
For ellipsis, we show inflection/VP scores.}\label{tab:scores_for_noise_rates}
\vspace{-2ex}
\end{table}

At training time, CADec uses either a translation sampled from the base model or a corrupted reference translation as the first-pass translation of the current sentence.
The purpose of using a corrupted reference instead of just sampling is to teach CADec to rely on the base translation and not to change it much. 
In this section, we discuss the importance of the proposed training strategy.

Results for different values of $p$ are given in Table~\ref{tab:scores_for_noise_rates}. All models have about the same BLEU, not statistically significantly different from the baseline, but they are quite different in terms of incorporating context. The denoising positively influences almost all tasks except for deixis, yielding the largest improvement on lexical cohesion.

\section{Additional Related Work}
 In concurrent work, \citet{xiong-coherence-delib-2018} also propose a two-pass context-aware translation model inspired by deliberation network. However, while they consider a symmetric data scenario where all available training data has document-level context, and train all components jointly on this data, we focus on an asymmetric scenario where we have a large amount of sentence-level data, used to train our first-pass model, and a smaller amount of document-level data, used to train our second-pass decoder, keeping the first-pass model fixed.

Automatic evaluation of the discourse phenomena we consider is challenging.
For lexical cohesion, \newcite{wong-kit:2012:EMNLP-CoNLL} count the ratio between the number of repeated and lexically similar content words over the total number of content words in a target document.
However, \newcite{guillou:2013:DiscoMT,carpuat-simard:2012:WMT} find that translations generated by a machine translation system tend to be similarly or more lexically consistent, as measured by a similar metric, than human ones. This even holds for sentence-level systems, where the increased consistency is not due to improved cohesion, but accidental -- \newcite{DBLP:conf/icml/OttAGR18} show that beam search introduces a bias towards frequent words, which could be one factor explaining this finding. This means that a higher repetition rate does not mean that a translation system is in fact more cohesive, and we find that even our baseline is more repetitive than the human reference.

\section{Conclusions}
We analyze which phenomena cause otherwise good context-agnostic translations to be inconsistent when placed in the context of each other. Our human study on an English--Russian dataset identifies deixis, ellipsis and lexical cohesion as three main sources of inconsistency. 
We create test sets focusing specifically on the identified phenomena. 

We consider a novel and realistic set-up where  a much larger amount of sentence-level data is available compared to that aligned at the document level and introduce a model suitable for this scenario. We show that our model effectively handles contextual phenomena without sacrificing general quality as measured with BLEU despite using only a small amount of document-level data, while a naive approach to combining sentence-level and document-level data leads to a drop in performance. We show that the proposed test sets allow us to distinguish models (even though identical in BLEU) in terms of their consistency.
To build context-aware machine translation systems, such targeted test sets should prove useful, for validation, early stopping and for model selection.

\section*{Acknowledgments}
We would like to thank the anonymous reviewers for their comments and Ekaterina Enikeeva for the help with initial phenomena classification.
The authors also thank Yandex Machine Translation team for helpful discussions and inspiration. Ivan Titov acknowledges support of the European Research Council (ERC StG BroadSem 678254) and the Dutch National Science Foundation (NWO VIDI 639.022.518). 
Rico Sennrich acknowledges support from the Swiss National Science Foundation (105212\_169888), the European Union’s Horizon 2020 research and innovation programme (grant agreement no 825460), and the Royal Society (NAF\textbackslash R1\textbackslash 180122). 

\nocite{training-tips-transformer}
\nocite{adam-optimizer}
\nocite{pymorphy}
\nocite{sennrich-bpe}

\bibliography{naaclhlt2019}
\bibliographystyle{acl_natbib}

\appendix

\section{Protocols for test sets}
\label{sec:appendix_testsets}
In this section we describe the process of constructing the test suites.
\subsection{Deixis}
\label{sec:appendix_testsets_deixis}
English second person pronoun ``you'' may have three different interpretations important
when translating into Russian: the second person singular informal (T form), the second person singular formal (V form) and second person plural (there is no T-V distinction for the plural from of second person pronouns).

Morphological forms for second person singular (V form) and second person plural pronoun are the same, that is why to automatically identify examples in the second person polite form, we look for morphological forms corresponding to second person plural pronouns.

To derive morphological tags for Russian, we use publicly available {\tt pymorphy2}\footnote{\url{https://github.com/kmike/pymorphy2}}~\cite{pymorphy}.

Below, all the steps performed to obtain the test suite are described in detail.

\subsubsection{Automatic identification of politeness}
For each sentence we try to automatically find indications of using T or V form. Presence of the following words and morphological forms are used as indication of usage of T/V forms:
\begin{enumerate}
    \item second person singular or plural pronoun,
    \item verb in a form corresponding to second person singular/plural pronoun,
    \item verbs in imperative form,
    \item possessive forms of second person pronouns.
\end{enumerate}

For 1-3 we used morphological tags predicted by {\tt pymorphy2}, for 4th we used hand-crafted lists of forms of second person pronouns, because {\tt pymorphy2} fails to identify them.

\subsubsection{Human postprocessing of identification of politeness}
After examples with presence of indication of usage of T/V form are extracted automatically, we manually filter out examples where
\begin{enumerate}
    \item second person plural form corresponds to plural pronoun, not V form,
    \item there is a clear indication of politeness.
\end{enumerate}
The first rule is needed as morphological forms for second person plural and second person singular V form pronouns and related verbs are the same, and there is no simple and reliable way to distinguish these two automatically.

The second rule is to exclude cases where there is only one appropriate level of politeness according to the relation between  the speaker and the listener. Such markers  include ``Mr.'', ``Mrs.'', ``officer'', ``your honour'' and ``sir''. For the impolite form, these include terms denoting family relationship (``mom'', ``dad''), terms of endearment (``honey'', ``sweetie'') and words like ``dude'' and  ``pal''.

\subsubsection{Automatic change of politeness}
To construct contrastive examples aiming to test the ability of a system to produce translations with consistent level of politeness, we have to produce an alternative translation by switching the formality of the reference translation. First, we do it automatically:
\begin{enumerate}
    \item change the grammatical number of second person pronouns, verbs, imperative verbs,
    \item change the grammatical number of possessive pronouns.
\end{enumerate}
For the first transformation we use {\tt pymorphy2}, for the second use manual lists of possessive second person pronouns, because {\tt pymorphy2} can not change them automatically.

\subsubsection{Human postprocessing of automatic change of politeness}
We manually correct the translations from the previous step. Mistakes of the described automatic change of politeness happen because of:
\begin{enumerate}
    \item ambiguity arising when imperative and indicative verb forms are the same,
    \item inability of {\tt pymorphy2} to inflect the singular number to some verb forms (e.g., to inflect singular number to past tense verbs), 
    \item presence of related adjectives, which have to agree with the pronoun,
    \item ambiguity arising when a plural form of a pronoun may have different singular forms.
\end{enumerate}

\subsubsection{Human annotation: are both polite and impolite versions appropriate?}
After the four previous steps, we have text fragments of several consecutive sentences with consistent level of politeness. Each fragment uses second person singular pronouns, either T form or V form, without nominal markers indicating which of the forms is the only one appropriate. For each group we have both the original version, and the version with the switched formality.

To control for appropriateness of both levels of politeness in the context of a whole text fragment we conduct a human annotation. Namely, humans are given both versions of the  same text fragment corresponding to different levels of politeness, and asked if these versions are natural. The answers they can pick are the following:
\begin{enumerate}
    \item both appropriate,
    \item polite version is not appropriate,
    \item impolite version is not appropriate,
    \item both versions are bad.
\end{enumerate}

The annotators are not given any specific guidelines, and asked to answer according to their intuition as a native speaker of the language (Russian).

There are a small number of examples where one of the versions is not appropriate and not equally natural as the other one: 4\%. Cases where annotators claimed both versions to be bad come from mistakes in target translations: OpenSubtitles data is not perfect, and target sides contain translations which are not reasonable sentences in Russian. These account for 1{.}5\% of all examples. We do not include these 5{.}5\% of examples in the resulting test sets.

\subsection{Lexical cohesion}
The process of creating the lexical cohesion test set consists of several stages:
\begin{enumerate}
    \item find passages where named entities are translated consistently,
    \item extract alternative translations for these named entities from the lexical table of Moses~\cite{moses} induced from the training data,
    \item construct alternative translations of each example by switching the translation of instances of the named entity,
    \item for each example construct several test instances. 
\end{enumerate}

\subsubsection{Identification of examples with consistent translations}
\label{identcohesion}

We look for infrequent words that are translated consistently in a text fragment. Since the target language has rich morphology, to verify that translations are the same we have to use lemmas of the translations. More precisely, we
\begin{enumerate}
    \item train Berkeley aligner on about 6{.}5m sentence pairs from both training and held-out data,
    \item find lemmas of all words in the reference translations in the held-out data using {\tt pymorphy2},
    \item find words in the source which are not in the 5000 most frequent words in our vocabulary whose translations have the same lemma.
\end{enumerate}

\subsubsection{Finding alternative translations}
For the words under consideration, we find alternative translations which would be (i) equally appropriate in the context of the remaining sentence and text fragment (ii) possible for the model to produce. To address the first point, we focus on named entities, and we assume that all translations of a given named entity seen in the training data are appropriate. To address the second point, we choose alternative translations from the reference translations encountered in the training data, and pick only ones with a probability at least 10\%. 

The sequence of actions is as follows:

\begin{enumerate}
    \item train Moses on the training data (6m sentence pairs),
    \item for each word under consideration (from \ref{identcohesion}), get possible translations from the lexical table of Moses,
    \item group possible translations by their lemma using {\tt pymorphy2},
    \item if a lemma has a probability at least 10\%, we consider this lemma as possible translation for the word under consideration,
    \item leave only examples with the word under consideration having several alternative translations.
\end{enumerate}

After that, more than 90\% of examples are translations of named entities (incl. names of  geographical objects). We manually filter the examples with named entities.

\subsubsection{Constructing a test set}
From the two previous steps, we have examples with named entities in context and source sentences and several alternative translations for each named entity. Then we
\begin{enumerate}
    \item construct alternative translations of each example by switching the translation of instances of the named entity; since the target language has rich morphology, we do it manually,
    \item for each example, construct several test instances. For each version of the translation of a named entity, we use this translation in the context, and vary the translation of the entity in the current sentence to create one consistent, and one or more inconsistent (contrastive) translation.
\end{enumerate}

\section{Experimental setup}

\subsection{Data preprocessing}
We use the publicly available OpenSubtitles2018 corpus~\cite{LISON18.294} for English and Russian.\footnote{\url{http://opus.nlpl.eu/OpenSubtitles2018.php}} 
We pick sentence pairs with a relative time overlap of subtitle frames between source and target language subtitles of at least $0.9$ to reduce noise in the data. As context, we take the previous sentence if its timestamp differs from the current one by no more than 7 seconds. Each long group of consecutive sentences is split into fragments of 4 sentences, with the first 3 sentences treated as context. More precisely, from a group of consecutive sentences $s_1, s_2, \dots, s_n$ we get $(s_1, \dots, s_4)$, $(s_2, \dots, s_5)$, $\dots$, $(s_{n-3}, s_{n})$. For CADec we also include $(s_1, s_2)$ and $(s_1, s_2, s_3)$ as training examples. We do not add these two groups with less context for the concatenation model, because in preliminary experiments, this performed worse both in terms of BLEU and consistency as measured on our test sets.

We use the tokenization provided by the corpus and use {\tt multi-bleu.perl}\footnote{\url{https://github.com/moses-smt/mosesdecoder/tree/master/scripts/generic}} on lowercased data to compute BLEU score. We use beam search with a beam of 4 for both base model and CADec.

Sentences were encoded using byte-pair encoding~\cite{sennrich-bpe}, with source and target vocabularies of about 32000 tokens.
Translation pairs were batched together by approximate sequence length. For the Transformer models (baselines and concatenation) each training batch contained a set of translation pairs containing approximately 16000\footnote{This can be reached by using several of GPUs or by accumulating the gradients for several batches and then making an update.} source tokens. It has been shown that Transformer's performance depends heavily on the batch size~\cite{training-tips-transformer}, and we chose a large batch size to ensure that models show their best performance. For CADec, we use a batch size that contains approximately the same number of translation instances as the baseline models.

\subsection{Model parameters}

We follow the setup of Transformer base model~\cite{attention-is-all-you-need}. More precisely, the number of layers in the base encoder, base decoder and CADed is $N=6$. We employ $h = 8$ parallel attention layers, or heads. The dimensionality of input and output is $d_{model} = 512$, and the inner-layer of a feed-forward networks has dimensionality $d_{ff}=2048$.

We use regularization as described in~\cite{attention-is-all-you-need}.

\subsection{Optimizer}
The optimizer we use is the same as in~\cite{attention-is-all-you-need}.
We use the Adam optimizer~\cite{adam-optimizer} with $\beta_1 = 0{.}9$, $\beta_2 = 0{.}98$ and $\varepsilon = 10^{-9}$. We vary the learning rate over the course of training, according to the formula:
\begin{multline*}
l_{rate}=scale\cdot \min(step\_num^{-0.5},\\ step\_num\cdot warmup\_steps^{-1.5}) 
\end{multline*}

We use $warmup\_steps = 16000$, $scale=4$ for the models trained on 6m data (baseline (6m) and concatenation) and $scale=1$ for the models trained on 1{.}5m data (baseline (1{.}5m) and CADec).

\end{document}